\documentclass{article}

\PassOptionsToPackage{numbers, compress}{natbib}
\usepackage[final]{neurips_2020_ml4ps}

\usepackage[utf8]{inputenc} % allow utf-8 input
\usepackage[T1]{fontenc}    % use 8-bit T1 fonts
\usepackage{hyperref}       % hyperlinks
\usepackage{url}            % simple URL typesetting
\usepackage{booktabs}       % professional-quality tables
\usepackage{amsfonts}       % blackboard math symbols
\usepackage{nicefrac}       % compact symbols for 1/2, etc.
\usepackage{microtype}      % microtypography
\usepackage{comment}

\usepackage{graphicx}
\usepackage{subcaption}
\usepackage{algpseudocode}
\usepackage{xspace}
\usepackage{tabu}
\usepackage{bm}
\usepackage{multirow}
\usepackage{enumitem}
\usepackage[ruled,vlined]{algorithm2e}
\usepackage{color,soul}
\usepackage{xcolor}
\usepackage{wrapfig}

\usepackage{amsmath}

\title{HydroNet: Benchmark Tasks for Preserving Intermolecular Interactions and Structural Motifs in Predictive and Generative Models for Molecular Data}

\author{%
  Sutanay Choudhury \\
  \texttt{sutanay.choudhury@pnnl.gov} \\
  \And
  Jenna A. Bilbrey \\
  \texttt{jenna.bilbrey@pnnl.gov} \\
  \And
  Logan Ward \\
  \texttt{lward@anl.gov} \\
  \And
  Sotiris S. Xantheas \\
  \texttt{sotiris.xantheas@pnnl.gov} \\
  \And
  Ian Foster \\
  \texttt{foster@anl.gov} \\
  \And
  Joseph P. Heindel \\
  \texttt{heindelj@uw.edu} \\
  \And
  Ben Blaiszik \\
  \texttt{blaiszik@uchicago.edu} \\
  \And
  Marcus E. Schwarting \\
  \texttt{meschw04@uchicago.edu} \\
}

\usepackage{natbib}
\usepackage{graphicx}

\newif\iffinal
\finaltrue

\iffinal
  \newcommand\sutanay[1]{}
  \newcommand\logan[1]{}
  \newcommand\jenna[1]{}
  \newcommand\sotiris[1]{}
  \newcommand\marcus[1]{}
  \newcommand\ian[1]{}
  \newcommand\ben[1]{}

\else
  \newcommand\sutanay[1]{{\color{blue}[Sutanay: #1]}}
  \newcommand\logan[1]{{\color{violet}[Logan: #1]}}
  \newcommand\jenna[1]{{\color{green}[Jenna: #1]}}
  \newcommand\sotiris[1]{{\color{red}[Sotiris: #1]}}
  \newcommand\marcus[1]{{\color{cyan}[Marcus: #1]}}
  \newcommand\ian[1]{{\color{orange}[Ian: #1]}}
  \newcommand\ben[1]{{\color{blue}[Ben: #1]}}
\fi

\begin{document}

\maketitle

\begin{abstract}
%Change first sentence
Intermolecular and long-range interactions are central to phenomena as diverse as gene regulation, topological states of quantum materials, electrolyte transport in batteries, and the universal solvation properties of water. We present a set of challenge problems \footnote{https://exalearn.github.io/hydronet} for preserving intermolecular interactions and structural motifs in machine-learning approaches to chemical problems, through the use of a recently published dataset of 4.95 million water clusters held together by hydrogen bonding interactions and resulting in longer range structural patterns. The dataset provides spatial coordinates as well as two types of graph representations, to accommodate a variety of machine-learning practices.
\end{abstract}

\section{Introduction}

The application of machine-learning (ML) techniques such as supervised learning and generative models in chemistry is an active research area. ML-driven prediction of chemical properties and generation of molecular structures with tailored properties have emerged as attractive alternatives to expensive computational methods \cite{SchNetPack2019, smith2017ani, n2p2MPI2019, MolDQN, GCPN, de2018molgan, khemchandani2020deepgraphmol, madhawa2019graphnvp, shi2020graphaf}. 
Though increasingly used, graph representations of molecules often do not explicitly include non-covalent interactions such as hydrogen bonding, which poses difficulties when examining systems with intermolecular and/or long-range interactions \cite{gebauer2019symmetry}.
To facilitate the development of such methods, we discuss a set of challenge problems and suggest an approach based on a recently published database of low-energy water cluster minima lying within 5 kcal/mol from the putative minimum of each cluster size \cite{rakshit2019database}.  

% Our dataset, challenge problems along with codebase for modeling and evaluation is available at \url{}.

\textbf{Scientific motivation:} A water cluster is a discrete hydrogen bonded network of water molecules. While most interactions are short-range (i.e., between neighboring molecules) \cite{SCHAEFFER2008359}, there also exist substantial (\textasciitilde 20\%) many-body, longer-range interactions (i.e., with next-nearest and more distant neighbors) \cite{Homodromic2000}. Understanding many-body and long-range hydrogen bonding interactions is key to answering long-standing scientific questions such as the macroscopic properties of liquid water, ice, and aqueous systems (e.g., heat capacity, density, dielectric constant, compressibility) \cite{keutsch2001water}. These interactions are responsible for the bulk and interfacial properties of liquid water and ice, as well as solvation processes, and are key to the realization of diverse applications, from drug delivery to protein folding and the design of quantum materials and novel electrolytes for batteries \cite{lieberman2009comprehensive, veljkovic2011role, french2010long}.

\begin{wrapfigure}[11]{r}{0.5\textwidth}
  \vspace*{-0mm}
  \includegraphics[width=0.5\textwidth]{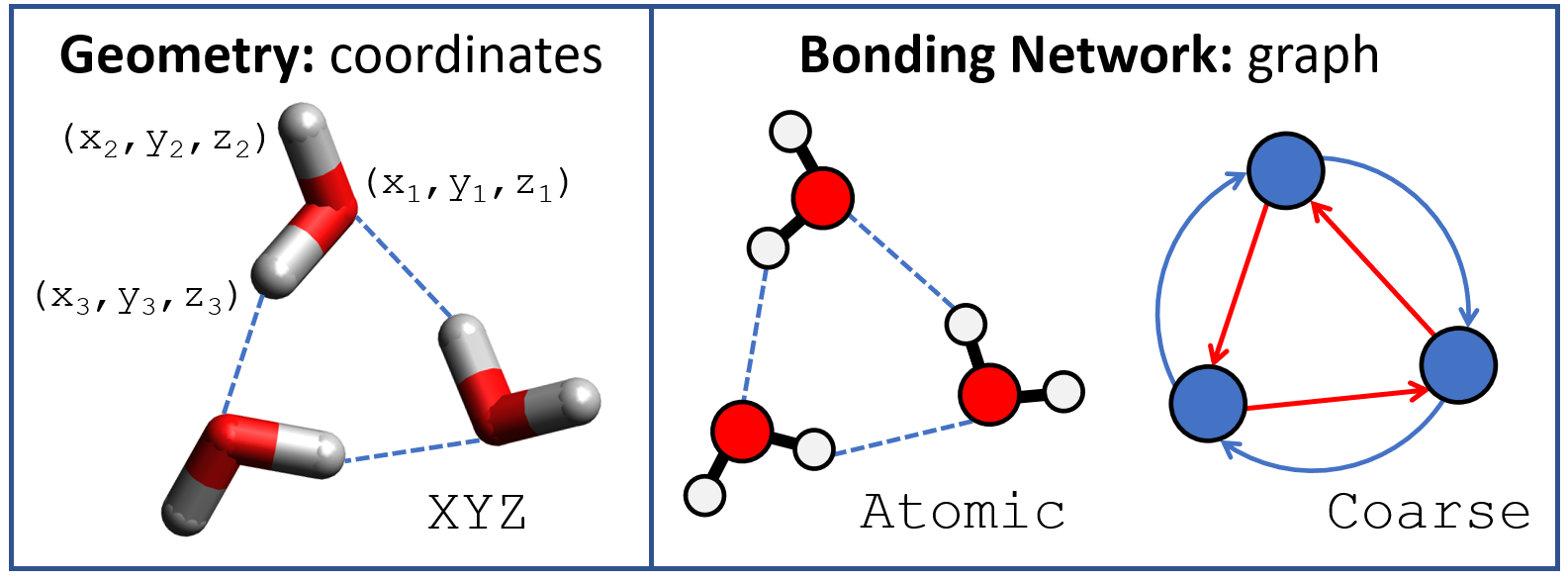}
  \vspace{-7mm}
  \caption{Data model: geometries provided as spatial coordinates.  In the graph representation nodes are atoms (Atomic) or water molecules (Coarse) and edges are covalent and/or hydrogen bonds.} \label{fig:dataset_types}
 \end{wrapfigure}
\textbf{HydroNet summary:} The dataset of 4.95 million water cluster minima is the largest collection of water cluster minima reported to date \cite{rakshit2019database}. Originally created to advance the development of interaction potentials in chemical physics, the dataset is composed of clusters of isomers differing in the underlying hydrogen bonding network. 
Each cluster is described by its potential energy, Cartesian coordinates of each atom, and two graph-based representations of their bonding arrangement (Fig.~\ref{fig:dataset_types}): 
an \emph{atomic interaction graph} that captures both intramolecular and intermolecular bonding patterns and a \emph{coarse graph} that captures only the intermolecular structure of the cluster.

\textbf{ML tasks on the dataset:} 
1) \textit{Molecular Property Prediction}: given a water cluster with specified spatial coordinate information or bonding structure, predict its energy. 2) \textit{Molecule Generation}: given \emph{N} water molecules, generate candidate structures that conform to structural measures of low-energy configurations described in the key challenges below.

\begin{figure}[htbp] \centering
\vspace*{-3mm}
  \includegraphics[width=0.99\textwidth]{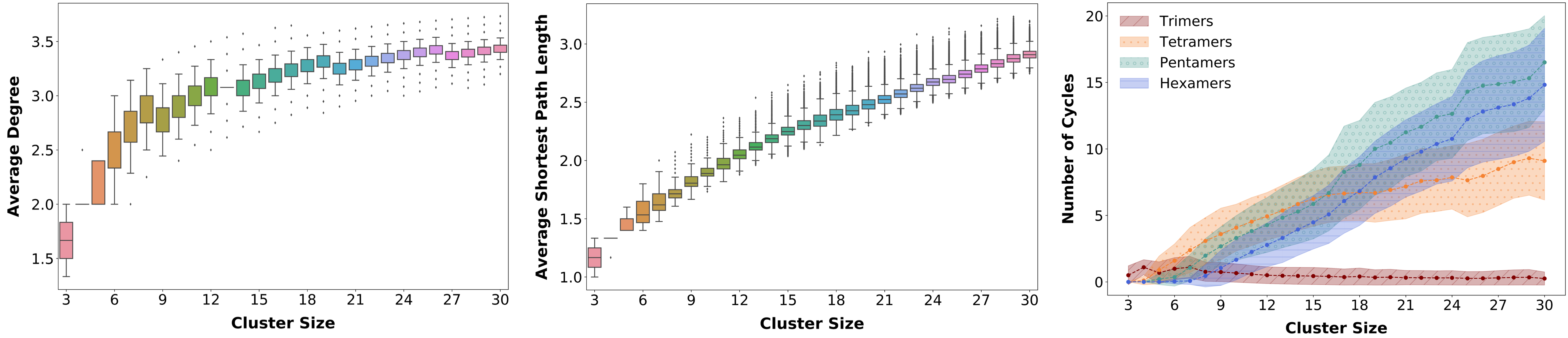}
  \caption{Graph-based measures capturing intermolecular interactions and structural motifs in water cluster networks. Clear patterns such as the average number of neighbors (left), the shortest path between two water molecules (middle), and the predominance of pentamer and hexamer polygons over tetramer polygons (right) emerge as the cluster size increases \cite{bilbrey2020jcp}.} \label{fig:distributions}
  \vspace*{-2mm}
\end{figure}

\textbf{Key challenges:} A defining feature of water clusters is that numerous dissimilar structures can have quite similar energies.  
In addition, for a given set of spatial orientations of oxygen atoms in the cluster (oxygen network), there exist numerous hydrogen bonding networks (Coarse graphs) depending on the arrangement of the hydrogen atoms, which form hydrogen bonds according to the Bernal--Fowler rules \cite{bernal1933}. 
The structural properties of low-energy hydrogen bonding networks -- characterized by graph-theoretical measures such as degree distribution, shortest path length, and distribution of polygons (Fig.~\ref{fig:distributions}) -- vary systematically with cluster size~\cite{bilbrey2020jcp}.
A generative method that produces a water cluster network should be mindful of these properties, as clusters far outside the distributions are likely to be much higher in energy and therefore of less interest. 

\section{Dataset Description}

The potential energy of each cluster was obtained using the ab initio-based Thole-type, flexible, polarizable interaction potential for water (TTM2.1-F) \cite{burnham2002development,fanourgakis2006flexible}.
Clusters were generated with the Monte Carlo Temperature Basin Paving (MCTBP) sampling method to produce a dense sampling of low-energy water clusters containing 3--30 water molecules per cluster \cite{rakshit2019database}. The water cluster minima dataset~\cite{hydrodb} is represented in the three formats shown in Fig.~\ref{fig:dataset_types}, each as both line-delimited JavaScript Object Notation (JSON) and Tensorflow Protobufs. Table~\ref{tab:data_formats} lists the information available for each sample.
The records are separated into predefined train (80\%), validation (10\%), and test sets (10\%), where clusters are maintained in the same subset for each data format. 
We also provide code to compute graph descriptors for structural motif tracking \cite{hydrocode}.

\begin{table}[htpb!]
    \centering
    \caption{Records in HydroNet are stored as dictionary objects with the below keys.}
    \label{tab:data_formats}
    \begin{tabular}{|p{1.7cm}|p{2.75cm}|p{8.21cm}|}
        \hline
        \textbf{Key} & \textbf{Type} & \textbf{Description} \\
        \hline
        \hline
        energy & float & Energy of the cluster, in kcal/mol. \\
        \hline
        n\_water & int & Number of water molecules. \\
        \hline
        n\_atom & int & Number of atoms or, for coarse graphs, nodes. \\
        \hline
        atom & list of ints & Types of each atom or node in the graph. Atomic: 0 is oxygen, 1 is hydrogen; coarse: all nodes are the same type. \\
        \hline
        z & list of ints & \textit{Geometry only}. Atomic number of each atom. \\
        \hline
        coords & $N\times3$ array of floats & \textit{Geometry only}. XYZ coordinates of atoms, in angstrom.\\
        \hline 
        n\_bond & list of ints & \textit{Graph only}. Number of edges in the graph. \\
        \hline 
        bond & list of ints & \textit{Graph only}. Types of edges. Atomic: 0 is covalent, 1 is hydrogen; coarse: 0 is donor, 1 is acceptor. \\
        \hline
        connectivity & $N\times2$ array of ints & \textit{Graph only}. Connectivity between nodes in the graph. \\
        \hline
    \end{tabular}
    
\end{table}

\section{Machine-Learning Tasks}\label{sec:tasks}

We introduce the defined property prediction and generative modeling tasks on the water cluster dataset and provide details on baseline implementations for the first task.

\subsection{Cluster Potential Energy Prediction Task}

This task is to predict the potential energy of a water cluster without the use of expensive ab initio methods, in two settings. 
In the \textit{geometry-to-energy} \cite{SchNet2018} setting, we assume the geometry of the structure is known accurately. In the \textit{graph-to-energy} \cite{gilmer2017neural, st2019message} setting, we require predictions to be made from the connectivity of the water molecules only.

We approach these tasks using neural network potentials, in which the original state of each atom ($h_v$) and bond ($\alpha_{vw}$) is represented as a vector embedding based on atomic number and bond type.  These states are modified by successive message layers.  Each message layer uses a multi-layer perceptron to compute a message from the atom state ($h_v$), a neighboring atom state ($h_w$), and the connecting bond ($\alpha_{vw}$). The atom and bond states are updated according to the following equations, which generalize both the geometry-to-energy and graph-to-energy prediction settings described below. In the first, an atom's neighbors are those within a radial cutoff distance; in the second, neighbors are represented by the adjacency matrix of the network.

\begin{comment}
\begin{equation*}
m^{t+1}_v = \sum_{w \in Neighbors(v)} M_t(h^t_v, h^t_w, \alpha^{t}_{vw})
\end{equation*}
\begin{equation*}
h^{t+1}_v = h^t_v + m^{t+1}_v
\end{equation*}
\begin{equation*}
\alpha^{t+1}_{vw} = \alpha^t_{vw} + M_t(h^t_v, h^t_w, \alpha^{t}_{vw})
\end{equation*}
\end{comment}

\begin{equation*}
m^{t+1}_v = \sum_{w \in \emph{Nbors}(v)} M_t(h^t_v, h^t_w, \alpha^{t}_{vw})\ \ \ \ \ \ \ \ \  h^{t+1}_v = h^t_v + m^{t+1}_v\ \ \ \ \ \ \ \ \ \alpha^{t+1}_{vw} = \alpha^t_{vw} + M_t(h^t_v, h^t_w, \alpha^{t}_{vw})
\end{equation*}

\begin{wrapfigure}[23]{r}{0.5\textwidth}
  %\vspace*{-4mm}
  \includegraphics[width=0.5\textwidth]{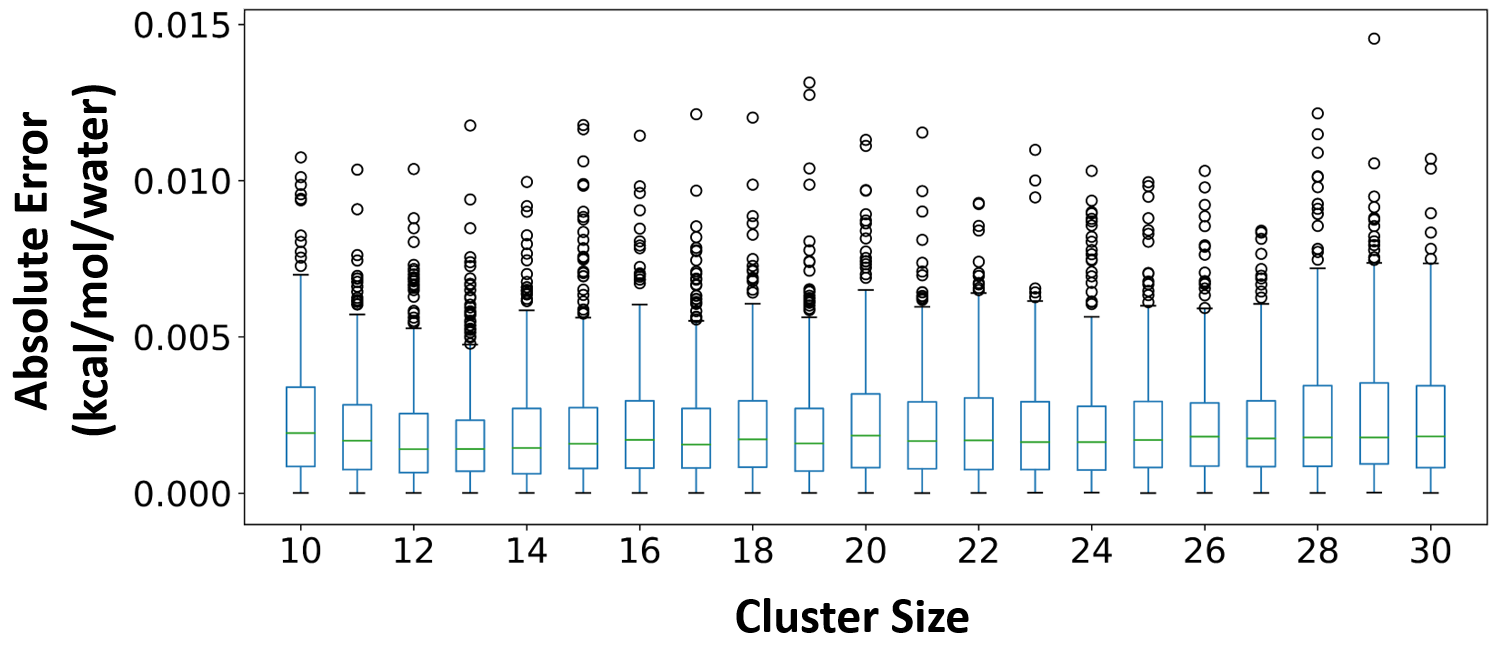}
  \includegraphics[width=0.5\textwidth]{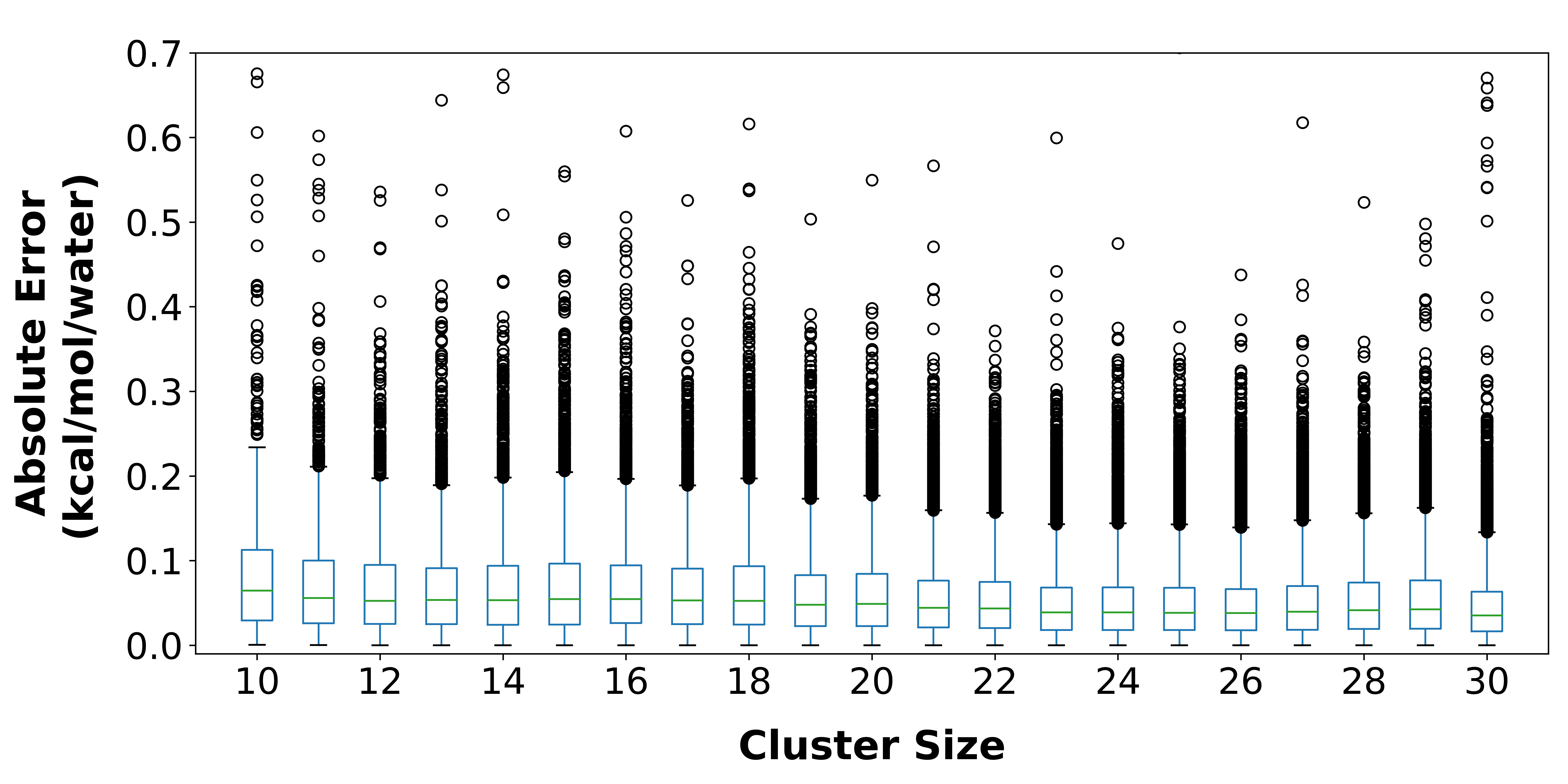}
  \vspace{-5mm}
  \caption{Performance of (top) SchNet trained on 500,000 water clusters, \emph{N}=11--29, and tested on 10,500 clusters, \emph{N}=10--30, and (bottom) MPNN trained on HydroNet training set and evaluated on test set. SchNet performance data from~\cite{bilbrey2020jcp}.} \label{fig:schnet_boxplot}
\end{wrapfigure}

We sum the energy per atom values produced by the penultimate layer to obtain the full cluster energy.

For the geometry-to-energy problem, we previously \cite{bilbrey2020jcp} trained a continuous-filter convolutional neural network (SchNet, PyTorch implementation \cite{SchNetPack2019}) on 500,000 water clusters of size \emph{N}=11--29. The training set was stratified by cluster size, from which 10\% were reserved for validation during training. The test set was composed of 10,500 clusters, with 500 from each cluster size \textit{N}=10--30. Clusters of size \emph{N}=10, 30 allowed us to examine the effects of energy prediction on clusters smaller and larger than those in the training set. Using mean squared error (MSE) as the loss, we achieved a final training loss of 0.0030 (kcal/mol)$^2$ and final validation loss of 0.0035 (kcal/mol)$^2$. The mean absolute error (MAE) of the test set predictions was 0.0427 kcal/mol. The network took $\sim$65 hours to train using 4 NVIDIA V100 GPUs.

For the graph-to-energy problem, we use a message-passing neural network (MPNN) inspired by \cite{gilmer2017neural} to create a reference implementation~\cite{st2019message}.
A first performance study shows that models trained on coarse networks achieve losses an order of magnitude superior to those trained on atomic networks, but exhibit significantly worse performance on the test set.  Such generalization issues suggest opportunities to explore regularization techniques or new network designs to take advantage of coarse graphs.

%% TODO (WardLT): Bring out the "challenge problem" nature of this problem
Our best models have a MAE of $\sim$0.5~kcal/mol/water. 
Errors are nearly equal for clusters of sizes 10--30, 
giving us confidence that the MPNN is capturing the changes
in favored geometries as clusters increase in size.
However, these errors are around a factor of 100 higher than those achieved 
with SchNet, which incorporates knowledge of the atomic coordinates.
In short, we find conventional MPNN approaches to be promising solutions for energy prediction, but significant work is needed by the ML community to find models suitable for evaluating systems with intermolecular and/or long-range interactions.

\subsection{Structural Measure-Preserving Molecular Generation Task}

\begin{wrapfigure}[16]{r}{0.45\textwidth}
   \vspace*{-7mm}
  \begin{center}
    \includegraphics[width=0.45\textwidth,trim=5mm 1mm 2mm 5mm,clip]{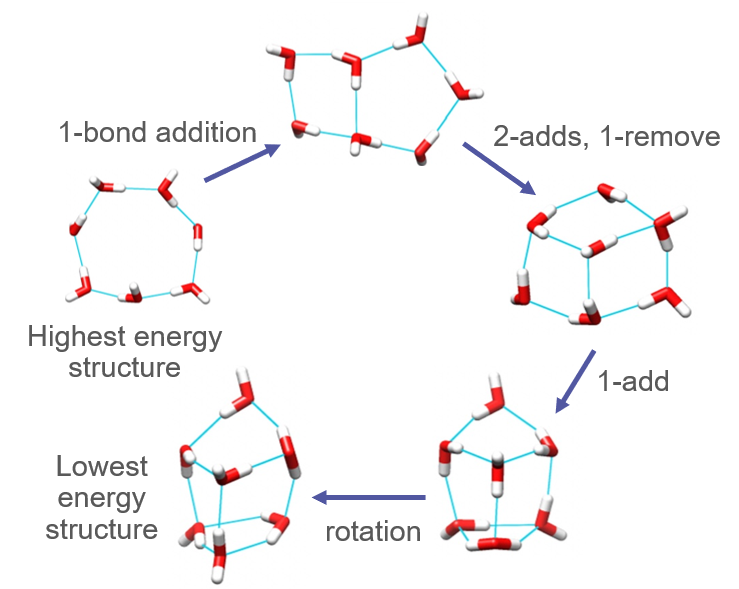}
  \end{center}
  \vspace*{-2mm}
  \caption{Generation of a low-energy water network from a higher energy structure.}
  \label{fig:waterdesign}
\end{wrapfigure}

We define the task as follows: given \emph{N} water molecules, generate a geometric/atomic/coarse representation that a) satisfies certain graph-theoretic structural measures derived from its hydrogen bonding network and b) minimizes the cluster energy by optimizing the relative spatial arrangements of atoms and molecules. 
As an example, Fig.~\ref{fig:waterdesign} shows a sequence of transformations that would yield a more stable water cluster from a static number of molecules.

\begin{figure*}[!h] \centering
  \includegraphics[width=0.95\textwidth]{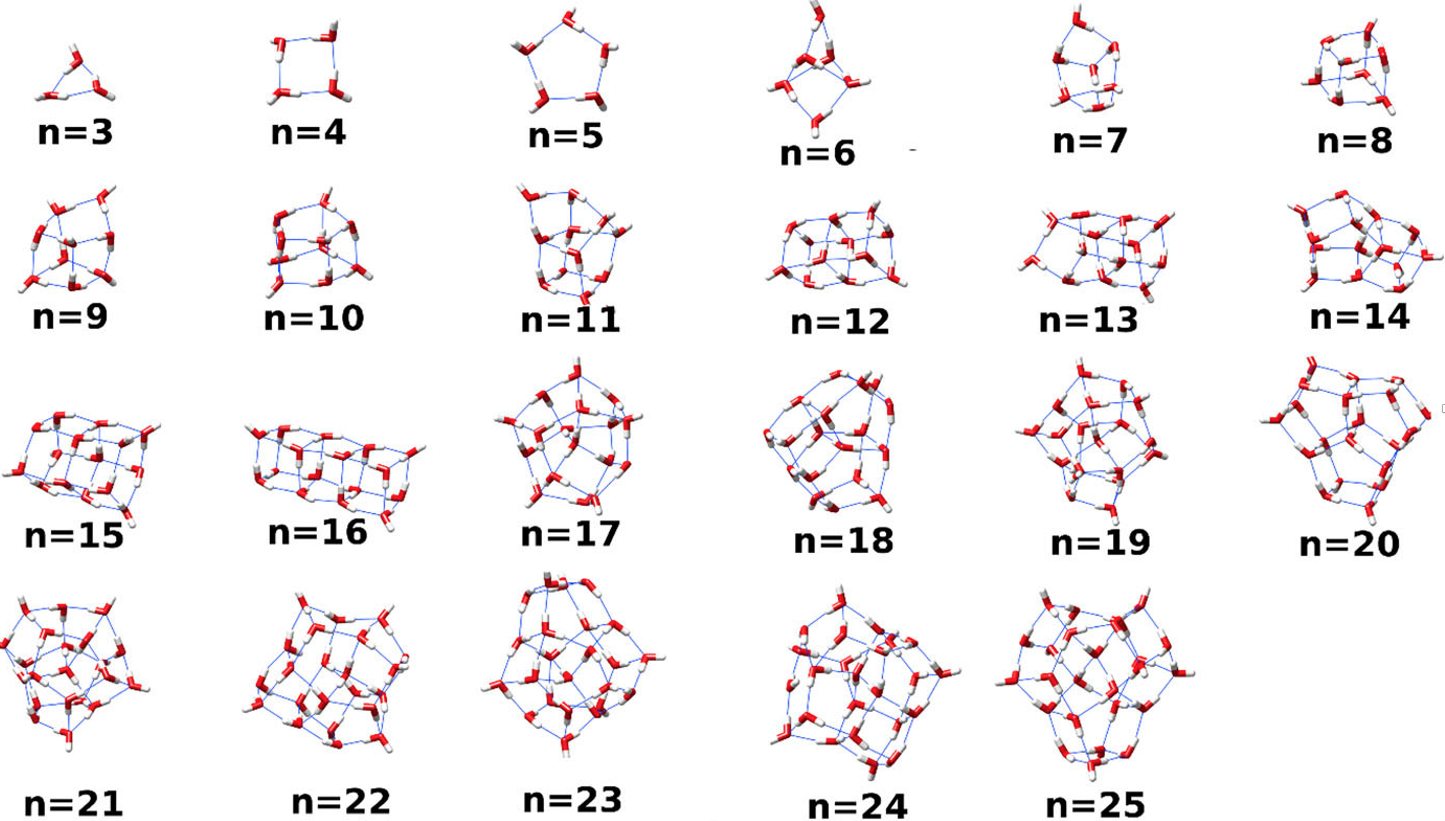}
  \caption{Geometric shape progression of low-energy structures with cluster size \emph{n}. Figure reproduced with permission from \cite{rakshit2019database}.} \label{fig:cluster_gallery}
\end{figure*}

Structural measures vary with cluster size, as shown in Fig.~\ref{fig:distributions}. The mean degree (number of neighbors) of water molecules in a hydrogen bonded network increases and eventually plateaus when water molecules become saturated with hydrogen bonds. The shortest path length also increases with cluster size, through the growth rate decays. 
The number of cycles in the hydrogen bonded network also show trends as clusters grow in size:  trimers (3-cycles) and tetramers (4-cycles) become less dominant, while pentamers (5-cycles) and hexamers (6-cycles) emerge as key building blocks.  
%We compute the number of cycles in the hydrogen network for different numbers of bonds and find that for larger clusters,  trimers (3-cycles) and tetramers (4-cycles) are less dominant, and pentamers (5-cycles) and hexamers (6-cycles) emerge as key building blocks as the cluster grows in size.  
We also observe a strong evolution in geometric structure (Fig.~\ref{fig:cluster_gallery}).  The water cluster structures start as quasi-planar graphs (\emph{N}=3-5), evolve into cubic graphs (\emph{N}=7–16), and then into cage-like structures (\emph{N}=17 onward), with some structures being quasi-symmetric (i.e. for \emph{N}=20).

\section{Relationship with Other Work}

% \begin{wrapfigure}[13]{r}{0.45\textwidth}
% \centering
%   \vspace*{-18mm}
%   \includegraphics[width=0.45\textwidth]{images/clusters-shapes.png}
%   \vspace{-6mm}
%   \caption{Geometric shape progression of low-energy structures with cluster size \emph{N}.} 
%   \label{fig:cluster_gallery}
% \end{wrapfigure}

Generative models and deep reinforcement learning methods that produce molecular structures with desired properties are active research areas \cite{GCPN, MolDQN, de2018molgan, jin2020composing, madhawa2019graphnvp, shi2020graphaf}.  However, most ML-benchmark datasets \cite{ramakrishnan2014quantum,sterling2015zinc,wu2018moleculenet} used by these studies map from spatial or graph structure to properties of interest (e.g, drug-likeliness, binding affinity between a drug molecule and a target protein). Modeling the interaction of heterogeneous functional groups and reasoning about potential reaction mechanisms is the primary focus for such tasks.  Our dataset and science challenge tasks require methods that prioritize satisfying topological constraints, addressing challenges with stereoisomers, and modeling variations in interaction patterns at multiple scales. 

The multi-scale nature of the interactions of our dataset also fills a void in existing scientific datasets.
Datasets which map molecular geometry or molecular graphs, available for years, capture challenges in which properties are driven by short-range covalent interactions \cite{smith2017anidata, wu2018moleculenet, ramakrishnan2014quantum}. 
There are also datasets of condensed phases (e.g., AGNI
\cite{Botu2016agni}) in which long-range, many-body interactions become important.
Our dataset presents a complex mixture of short-range covalent, intermolecular hydrogen bonding, and extended many-body effects.
We propose this dataset as a basis for the challenges of learning such effects from 3D geometries and inferring them from bonding structure.

\section{Conclusions}
We present challenge problems and a corresponding dataset to advance machine-learning on molecular data.
An important feature of the dataset discussed here is the incorporation of intermolecular interactions, which allows for two distinct graph representations in addition to the spatial coordinates. 
Our initial benchmarks illustrate a large performance gap between neural networks based on atomic coordinates and those which use only the bonding structure, which make this benchmark particularly relevant for advancing graph neural network-based generative and prediction models.
Moreover, this dataset is useful for the creation of novel machine-learning methods that preserve structural patterns across a wide size regime and can address the key challenges of property prediction and molecular generation.

\section{Broader Impact}
The problems addressed here have the potential to advance methods for modeling biological and chemical systems, leading to vast speed-ups in computation that would accelerate accurate simulations of large, complex systems. These speed-ups will allow domain researchers to more quickly assess the properties of and generate potential structures for novel molecular systems with phenomena dependant on long-range interactions. Such research is the beginning stage to making practicable, real-world scientific advancements.

\section{Acknowledgements}
%Line for Argonne
S.C., J.A.B., L.W., S.S.X., I.F., J.P.H., and M.S. were supported by the DOE Exascale Computing Project, ExaLearn Co-design Center. B.B. was was supported by the National Science Foundation under NSF Award Number: 1931306 ``Collaborative Research: Framework: Machine Learning Materials Innovation Infrastructure.'' This research used resources of the Argonne Leadership Computing Facility, which is a DOE Office of Science User Facility supported under Contract DE-AC02-06CH11357. Pacific Northwest National Laboratory (PNNL) is a multi-program national laboratory operated for the US Department of Energy by Battelle. Argonne National Laboratory's work was supported by the U.S. Department of Energy, Assistant Secretary for Environmental Management, Office of Science and Technology, under contract DE-AC02-06CH11357.
%Line for Argonne support

%\section{Acknowledgments and Disclosure of Funding}

\bibliographystyle{plain}
\bibliography{references}
\end{document}